\documentclass[11pt]{article}
\usepackage{amssymb}
\usepackage[preprint]{acl}

\usepackage{times}
\usepackage{latexsym}

\usepackage[T1]{fontenc}
\usepackage[most]{tcolorbox}
\tcbuselibrary{listingsutf8}  

\definecolor{mygray}{gray}{.9}
\definecolor{c0}{cmyk}{1,0.3968,0,0.2588} 
\tcbset{
  colback=blue!5!white,
  colback=blue!3!white,
  colbacklower=blue!10!white,
  colframe=blue!40!black,
  boxrule=1.2pt,
  arc=5pt,
  left=4pt, right=4pt, top=4pt, bottom=4pt,
  fonttitle=\bfseries,
  fontupper=\ttfamily,
  enhanced,
  coltitle=white,
  drop fuzzy shadow,
  borderline={0.6pt}{0pt}{blue!55!black}
}

\usepackage[utf8]{inputenc}

\usepackage{microtype}

\usepackage{inconsolata}

\usepackage{graphicx}
\usepackage{algorithm}
\usepackage{algorithmic}
\usepackage{amsfonts}
\usepackage{tcolorbox}
\usepackage{pifont}
\tcbuselibrary{breakable} 
\tcbuselibrary{skins} 
\usepackage{enumitem}
\usepackage{xcolor}
\usepackage{colortbl}
\usepackage{amsmath}
\definecolor{lightblue}{RGB}{200,220,240}  
\definecolor{lighterlightblue}{RGB}{235,240,252}  

\usepackage{algorithm}
\usepackage{algorithmic}
\usepackage{multirow}
\usepackage{booktabs}
\usepackage{tabularx}
\usepackage{array}
\title{Efficient Reasoning via Thought-Training and Thought-Free Inference}

\author{%
    Canhui Wu$^{1,2}$\thanks{Work completed during internship at JD.com}, Qiong Cao$^{2}$\thanks{Corresponding author},  Chao Xue$^{2}$, \textbf{Wei Xi}$^{1}$, \textbf{Xiaodong He}$^{2}$ \vspace{0.5em}
    \\
    $^{1}$Xi'an Jiaotong University,  \hspace{0.3em}$^{2}$JD Future Academy \\
    \texttt{wucanhui@stu.xjtu.edu.cn} \quad \texttt{caoqiong1@jd.com}
}

\begin{document}
\maketitle
\begin{abstract}
Recent advances in large language models (LLMs) have leveraged explicit Chain-of-Thought (CoT) prompting to improve reasoning accuracy. However, most existing methods primarily focus on compressing verbose reasoning outputs. These Long-to-Short transformations aim to improve efficiency, but require a large amount of short CoT data. In this work, we introduce \textbf{3TF} (\textbf{T}hought-\textbf{T}raining and \textbf{T}hought-\textbf{F}ree inference), a framework for efficient reasoning that takes a Short-to-Long perspective. We first train a hybrid model that can operate in both reasoning and non-reasoning modes, and then further train it on CoT-annotated data to internalize structured reasoning, while enforcing concise, thought-free outputs at inference time using the no-reasoning mode. Unlike compression-based approaches, 3TF improves the reasoning quality of non-reasoning outputs, enabling models to perform rich internal reasoning implicitly while keeping external outputs short. Empirically, 3TF-trained models obtain large improvements on reasoning benchmarks under thought-free inference, demonstrating that high quality reasoning can be learned and executed implicitly without explicit step-by-step generation.
\end{abstract}

\section{Introduction}

Large language models (LLMs) have demonstrated remarkable reasoning abilities across arithmetic, commonsense, and symbolic domains. Chain-of-Thought (CoT) prompting~\cite{wei2022chain,kojima2022large} further enhances these abilities by encouraging models to decompose complex problems into intermediate reasoning steps. This explicit reasoning paradigm has proven crucial for improving model interpretability and success rates on multi-step reasoning tasks.

However, recent studies have revealed a growing inefficiency in the reasoning process of advanced LLMs. When solving simple problems such as “What is 2+3?”, models may produce unnecessarily lengthy explanations or enter cycles of self-reflection—a phenomenon known as \textit{overthinking}~\cite{sui2025stop,chen2024not}. Overthinking not only increases token consumption and latency~\cite{aytes2025sketch}, but also risks introducing compounding reasoning errors through redundant or self-contradictory steps~\cite{cuadron2025danger,su2025between}. As reasoning-oriented models continue to scale, mitigating this inefficiency has become a critical research focus.

To improve reasoning efficiency, existing approaches generally fall into three categories.
Prompt-based methods guide models to produce shorter responses through concise or controlled prompting~\cite{renze2024benefits,xu2025chain}. While simple and effective in some cases, these methods rely heavily on handcrafted instructions and often fail to generalize across tasks.
Reinforcement learning (RL)-based strategies offer a more adaptive solution by rewarding correctness while penalizing excessive token usage~\cite{luo2025o1,yi2025shorterbetter,arora2025traininglanguagemodelsreason,hou2025thinkprune}.
Supervised fine-tuning (SFT) approaches attempt to train models on datasets with concise reasoning traces, promoting brevity through direct supervision~\cite{ma2025cot,kang2025c3ot,xia2025tokenskip,yu2025long}.
All these methods can be viewed as Long-to-Short approaches, aiming to reduce reasoning length while maintaining performance.

In this work, we propose \textbf{3TF}, a new framework that approaches efficient reasoning from a Short-to-Long perspective. Instead of compressing explicit reasoning traces, 3TF employs an asymmetric training and inference paradigm. During training, the model learns from CoT-annotated data to internalize reasoning patterns. During inference, it operates in a thought-free mode, directly outputting concise final answers. This design allows the model to retain the benefits of CoT supervision, such as structured internal reasoning, while avoiding explicit and verbose outputs that lead to overthinking.

Previous SFT approaches typically require access to shortened CoT data for model training, such as summarization via LLMs, manual rewriting, or model merging, which can be costly and time-consuming. Unlike prior methods that explicitly constrain reasoning length, 3TF emphasizes implicit reasoning transfer. The training of 3TF is conducted in two stages: in the first stage, we train a hybrid reasoning model (HRM) capable of flexibly switching between explicit and implicit reasoning modes. In the second stage, the model learns exclusively from detailed CoT data to further strengthen its reasoning capability. The resulting model internalizes structured reasoning patterns, which can be silently activated during inference, effectively transforming reasoning from an explicit linguistic process into a latent cognitive one.

Extensive experiments show that 3TF consistently maintains high reasoning accuracy while substantially reducing output length. On GSM8K and MATH500, it retains over 90\% accuracy compared to full-CoT baselines while using roughly one third of the tokens; on harder benchmarks such as OLYMPIC and AIME24, it preserves 70--80\% of full-CoT performance with significantly shorter outputs. These results indicate that effective reasoning can be internalized and executed implicitly, without requiring explicit intermediate steps, highlighting that high-quality reasoning need not always be verbalized to be effective.

Our contributions are summarized as follows:
\begin{enumerate}[leftmargin=*]
\item We propose \textbf{3TF}, an asymmetric Short-to-Long framework that decouples training and inference, enabling LLMs to internalize reasoning via CoT supervision while producing concise, thought-free outputs.  
\item We show that reasoning-only training strengthens latent reasoning priors and improves implicit reasoning recoverability as model scale increases.  
\item We demonstrate that 3TF achieves state-of-the-art efficiency–accuracy trade-offs across multiple reasoning benchmarks, offering a new paradigm for efficient reasoning in LLMs.
\end{enumerate}
\section{Related Work}

\subsection{Reasoning in LLMs}

Large language models (LLMs) have demonstrated remarkable capabilities in solving complex reasoning tasks, particularly when guided by carefully designed prompting strategies. A key advancement in this domain is Chain-of-Thought (CoT) prompting~\cite{wei2022chain, kojima2022large, wang2022self}, which encourages models to generate intermediate reasoning steps before producing a final answer. Subsequent works have shown that CoT improves performance across various reasoning benchmarks. Techniques such as self-consistency decoding~\cite{wang2022self} and multi-step verification~\cite{creswell2022selection} further refine CoT by aggregating multiple reasoning paths, leading to more robust and accurate outputs. To enhance these abilities even further, leading LLMs such as OpenAI-O1~\cite{jaech2024openai}, DeepSeek-R1~\cite{guo2025deepseek}, and QwQ~\cite{team2025qwq} integrate advanced methods including reinforcement learning and multi-stage training. These strategies enable detailed multi-step reasoning, achieving state-of-the-art performance in fields such as advanced mathematics and competitive programming. Despite its success, CoT inference introduces substantial computational overhead because the reasoning traces must be explicitly generated at inference time, motivating research into more efficient approaches that retain high reasoning performance.

\subsection{Efficient Reasoning}

\begin{figure*}[tb]
    \centering
    \includegraphics[width=\textwidth]{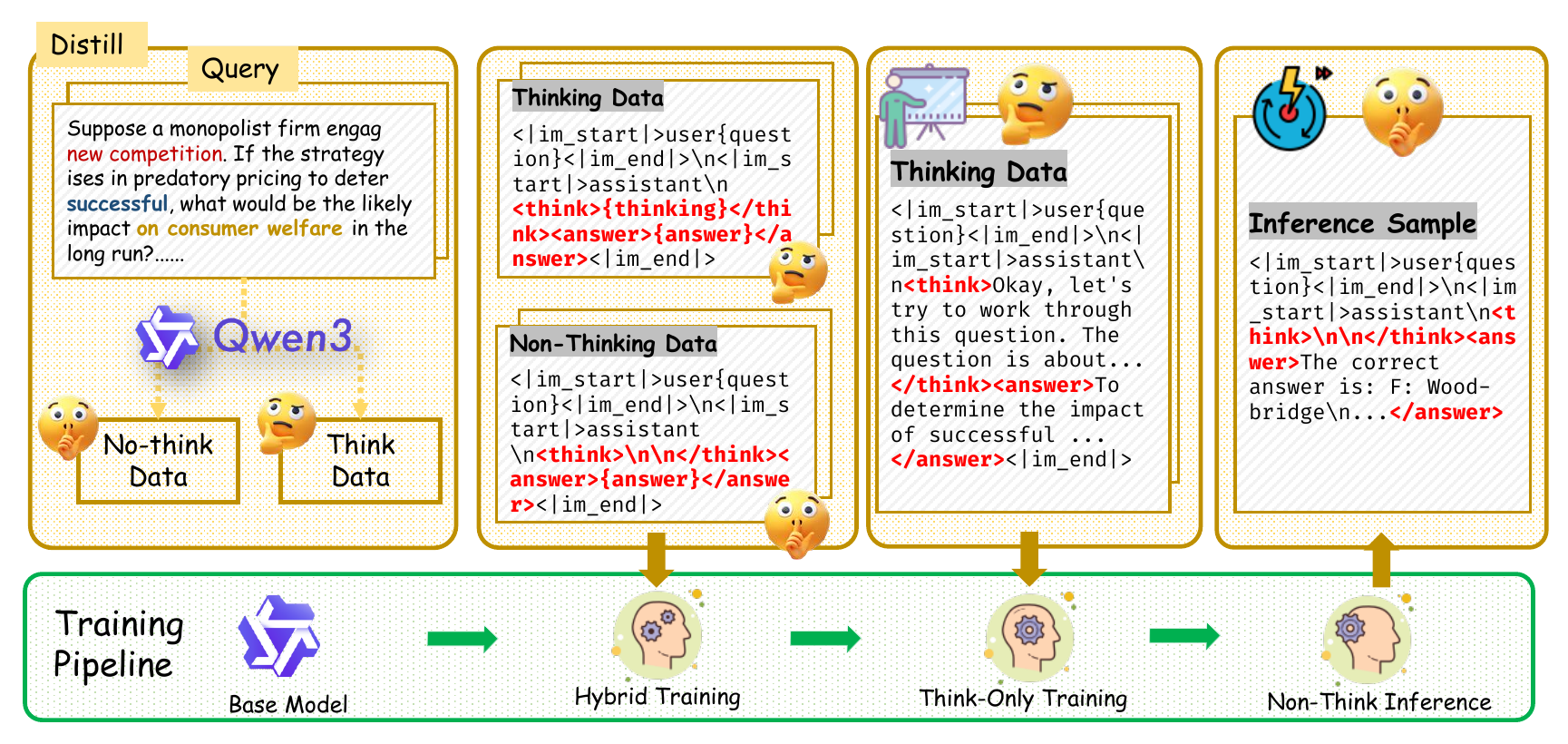}
    \caption{Overview of the 3TF Asymmetric Reasoning Paradigm.
\textbf{(Top)} Training and inference templates.
\textbf{(Bottom)} Flowchart: A \texttt{Base} model is trained on both \texttt{Think} and \texttt{No-Think} data, then fine-tuned on \texttt{Think} data. For \texttt{No-Think Inference}, the model uses the \texttt{Inference Template} to generate only the final answer.}

\label{fig:3tf}
\end{figure*}

As reasoning in LLMs becomes more verbose, recent works aim to make it more concise while preserving quality and accuracy. Prompt-based methods, such as CCoT~\cite{renze2024benefits}, guide models with explicit directives like ``Be concise,'' while CoD~\cite{xu2025chain} and Token-budget~\cite{han2024token} enforce strict token limits to prevent excessively long outputs. SFT-based approaches, including C3ot~\cite{kang2025c3ot}, Cot-Valve~\cite{ma2025cot}, TokenSkip~\cite{xia2025tokenskip}, and LS-Mixture~\cite{yu2025long}, fine-tune models on datasets containing reasoning of varying lengths, with particular emphasis on shorter and more efficient chains. Merge-based methods~\cite{wu2025unlocking} combine reasoning and instruction model parameters to reduce length efficiently without additional training, while RL-based techniques like DAST~\cite{shen2025dast}, O1-pruner~\cite{luo2025o1}, ShorterBetter~\cite{yi2025shorterbetter}, and TrainEfficient~\cite{arora2025traininglanguagemodelsreason} introduce length penalties during training to encourage concise outputs. Other work focuses on compressing reasoning, for example by distilling system 2 reasoning into system 1~\cite{yu2024distilling}, preserving reasoning capabilities while significantly reducing inference complexity.

\section{From Explicit Thoughts to Implicit Reasoning}
\label{sec:implicit_reasoning}

\subsection{Motivation and Intuition}

Recent hybrid reasoning models demonstrate the ability to alternate flexibly between explicit reasoning and direct answers, highlighting a growing interest in efficient reasoning that does not always require overt verbalization. However, explicit reasoning often remains verbose, unstable, and computationally expensive. To address this, we propose \textbf{3TF} (\textbf{T}hought-\textbf{T}raining and \textbf{T}hought-\textbf{F}ree inference), an asymmetric paradigm that decouples the reasoning process during training and inference. During training, the model learns from reasoning-augmented data, which includes both explicit reasoning steps and final answers, while at inference time, it is explicitly constrained to suppress reasoning and output only the final answer. This approach enables the model to benefit from reasoning supervision during training, while enforcing thought-free execution during deployment.

\subsection{Asymmetric Reasoning Process}

We denote each training instance as a triplet $(x, t, a)$, where $x$ represents the input (e.g., a question or problem description), $t$ denotes the reasoning trace that explicates the intermediate thought process, and $a$ corresponds to the final answer.
During training, the model learns to approximate the joint distribution of reasoning and answer generation conditioned on the input:
\begin{equation}
P_\theta(t, a \mid x) = P_\theta(t \mid x)\, P_\theta(a \mid t, x),
\label{eq:joint}
\end{equation}
where the first component $P_\theta(t \mid x)$ models the generation of reasoning steps, and the second component $P_\theta(a \mid t, x)$ models the derivation of the final answer given both the input and the reasoning trace.
This decomposition encourages the model to explicitly learn structured reasoning behavior, aligning its internal computation with interpretable chains of thought.

However, the inference stage deliberately adopts an asymmetric process.
Unlike training, where explicit reasoning tokens are available, inference suppresses their textual emission.
Formally, we denote the inference distribution as conditioning on an empty reasoning context:
\begin{equation}
P_\theta(a \mid x, t = \varnothing),
\label{eq:inference}
\end{equation}
where $t = \varnothing$ is not a random variable from the training distribution but an abstract symbol representing the fixed, empty reasoning context at inference.  
Although $P_\theta(a \mid x, t = \varnothing)$ is not directly optimized during training, the model’s behavior under this condition reflects its capacity for implicit reasoning.  

During deployment, decoding is initialized with an empty reasoning field such that the model is conditioned on $t = \varnothing$ from the outset.  
This simple structural constraint ensures that the model performs reasoning implicitly while emitting only the final answer, maintaining internal coherence without exposing intermediate steps.

Such an asymmetric design achieves two essential goals:  
(1) it retains the reasoning competence acquired through chain-of-thought supervision, and  
(2) it suppresses redundant or unstable intermediate outputs, improving inference efficiency and stability.  
Thus, 3TF provides controllable, interpretable, and computationally efficient reasoning behavior without requiring explicit traces at test time.

\subsection{Information-Theoretic Explanation}
\label{sec:info_theory_explanation}

From an information-theoretic view, explicit and implicit reasoning are complementary. Explicit reasoning expands input $x$ into intermediate traces $t$ ($x \rightarrow t \rightarrow a$), capturing fine-grained dependencies ($I(x; t) \gg I(x; a)$), while training aligns $t$ with $a$ via
\[
\max_\theta I_\theta(x; t) + I_\theta(t; a \mid x).
\]
At inference, suppressing $t$ compresses information directly into $a$, leveraging the aligned channels to retain essential semantics. Explicit reasoning expands and structures information; implicit reasoning compresses it efficiently.

\section{Experimental}\label{sec:exp}
\subsection{Experimental Setup}
\paragraph{Models and Datasets}
We conduct experiments on multiple sizes of the Qwen3~\cite{yang2025qwen3} model, including 4B, 8B, and 14B variants. Qwen3 is a hybrid reasoning model that is well-suited for experimental studies. For training, we randomly sample 100K problems from the MATH subset of AM-Qwen3-Distilled~\cite{tian2025correctanswersequaldistillation}. We evaluate 3TF on four reasoning benchmarks: AIME24, MATH500~\cite{lightman2023let}, GSM8K~\cite{cobbe2021gsm8k}, and Olympiad.

\begin{table*}[t]
\centering
\setlength{\tabcolsep}{1.4mm}
\renewcommand{\arraystretch}{1.05}
\begin{tabular}{@{}lrrcrrcrrcrrcr@{}}
\toprule
\multirow{2}{*}{\textbf{Methods}} & \multicolumn{3}{c}{\textbf{GSM8K@1}}  & \multicolumn{3}{c}{\textbf{MATH-500@4}} & \multicolumn{3}{c}{\textbf{OLYMPIC@4}} & \multicolumn{3}{c}{\textbf{AIME 2024@16}} \\ 
\cmidrule(lr){2-4} \cmidrule(lr){5-7} \cmidrule(lr){8-10} \cmidrule(lr){11-13}
&Acc. &Tok. &Ratio &Acc. &Tok. &Ratio &Acc. &Tok. &Ratio &Acc. &Tok. &Ratio \\ 
\midrule
\rowcolor{lightgray!30} \multicolumn{13}{c}{\textbf{\textit{Qwen3-4B}}} \\ 
\midrule
\texttt{Thinking} & 93.7 & 1792 & 100\% & 97.2 & 4229 & 100\% & 80.7 & 10122 & 100\% & 80.0 & 13200 & 100\%\\
\texttt{NoThinking} & 88.1 & 250 & 13.9\% & 91.6 & 972 & 23.0\% & 63.5 & 1778 & 17.6\% & 43.3 & 3792 & 28.7\%\\
\texttt{BeConcise} & 93.4 & 794 & 44.3\% & 97.8 & 3513 & 83.1\% & 80.7 & 8790 & 86.8\% & 80.0 & 13053 & 98.9\%\\
\texttt{ChainofDraft} & 94.3 & 567 & 31.6\% & 96.6 & 2457 & 58.1\% & 80.8 & 7339 & 72.5\% & 83.3 & 10626 & 80.5\%\\
\texttt{CoT-Valve} & 85.9 & 314 & 17.5\% & 91.4 & 1106 & 26.1\% & 61.8 & 1644 & 16.2\% & 53.3 & 1870 & 14.2\%\\
\texttt{LS-Mixture} & 93.1 & 1702 & 95.0\% & 96.2 & 3672 & 86.8\% & 78.5 & 8772 & 86.6\% & 73.3 & 12670 & 95.9\%\\
\rowcolor{c0!5} \textbf{3TF} & 92.4 & 278 & \textbf{15.5\%} & 95.6 & 1505 & \textbf{35.6\%} & 73.0 & 4097 & \textbf{40.5\%} & 70.0 & 9123 & \textbf{69.1\%}\\
\midrule
\rowcolor{lightgray!30} \multicolumn{13}{c}{\textbf{\textit{Qwen3-8B}}} \\ 
\midrule
\texttt{Thinking} & 94.2 & 2020 & 100\% & 98.0 & 4894 & 100\% & 81.4 & 10454 & 100\% & 80.0 & 13173 & 100\%\\
\texttt{NoThinking} & 89.9 & 245 & 12.1\% & 93.2 & 1096 & 22.4\% & 65.5 & 2081 & 19.9\% & 50.0 & 3581 & 27.2\%\\
\texttt{BeConcise} & 95.0 & 946 & 46.8\% & 98.8 & 4010 & 82.0\% & 81.6 & 9695 & 92.7\% & 83.3 & 12772 & 97.0\%\\
\texttt{ChainofDraft} & 94.0 & 500 & 24.8\% & 97.2 & 2978 & 60.8\% & 81.7 & 8386 & 80.2\% & 83.3 & 11365 & 86.3\%\\
\texttt{CoT-Valve} & 88.1 & 318 & 15.7\% & 93.2 & 764 & 15.6\% & 65.3 & 1934 & 18.5\% & 53.3 & 1916 & 14.5\%\\
\texttt{LS-Mixture} & 93.7 & 1563 & 77.4\% & 96.9 & 3735 & 76.3\% & 80.3 & 8382 & 80.2\% & 80.0 & 13091 & 99.4\%\\
\rowcolor{c0!5} \textbf{3TF} & 93.0 & 241 & \textbf{11.9\%} & 96.6 & 1358 & \textbf{27.7\%} & 76.0 & 3134 & \textbf{30.0\%} & 76.6 & 7961 & \textbf{60.4\%}\\
\midrule
\rowcolor{lightgray!30} \multicolumn{13}{c}{\textbf{\textit{Qwen3-14B}}} \\ 
\midrule
\texttt{Thinking} & 94.6 & 1525 & 100\% & 98.8 & 3748 & 100\% & 83.9 & 9675 & 100\% & 90.0 & 13253 & 100\%\\
\texttt{NoThinking} & 92.1 & 247 & 16.2\% & 94.4 & 803 & 21.4\% & 67.3 & 1791 & 18.5\% & 56.6 & 3153 & 23.8\%\\
\texttt{BeConcise} & 94.4 & 656 & 43.0\% & 99.0 & 3206 & 85.6\% & 83.0 & 8511 & 88.0\% & 90.0 & 12777 & 96.4\%\\
\texttt{ChainofDraft} & 95.3 & 585 & 38.4\% & 97.0 & 2286 & 61.0\% & 82.0 & 7006 & 72.4\% & 86.6 & 10441 & 78.7\%\\
\texttt{CoT-Valve} & 90.0 & 311 & 20.4\% & 94.0 & 814 & 21.7\% & 65.7 & 1510 & 15.6\% & 50.0 & 2105 & 15.9\%\\
\texttt{LS-Mixture} & 93.7 & 1248 & 81.8\% & 97.4 & 3281 & 87.5\% & 78.4 & 6788 & 70.1\% & 83.3 & 10584 & 79.8\%\\
\rowcolor{c0!5} \textbf{3TF} & 94.4 & 275 & \textbf{18.0\%} & 98.6 & 1613 & \textbf{43.0\%} & 79.4 & 4968 & \textbf{51.3\%} & 86.6 & 10280 & \textbf{77.5\%}\\
\bottomrule
\end{tabular}
\caption{Experimental results of 3TF on reasoning models of various sizes. For CoT-Valve and LS-mixture, we fine-tune the models on their official datasets.}
\label{tab:main}
\end{table*}

\paragraph{Implementation Details}
We fine-tune the Qwen3 model using the Swift~\cite{zhao2024swiftascalablelightweightinfrastructure} framework for 3 epochs with a batch size of 32. The learning rate is set to 2e-5 with a warmup ratio of 0.05. We enable sequence packing to improve training efficiency.  
We use LightEval~\cite{lighteval} for evaluation. The number of samplings during evaluation depends on the dataset: 16 samples per question for AIME 2024, 4 samples per question for MATH500 and Olympiad, and 1 sample per question for GSM8K. We compute pass@k based on these sampled generations.

\paragraph{Baselines}  

We compare our method against three types of baselines.  
1) \textbf{Qwen3 Modes}: The original Qwen3 model is evaluated under both the Think and No-Think modes to measure the impact of explicit reasoning instructions.  
2) \textbf{Prompt-based Responses}: We evaluate prompts that encourage the model to produce concise answers, including the BeConcise~\cite{renze2024benefits} and ChainofDraft~\cite{xu2025chain} strategies, which aim to streamline output while retaining correctness.  
3) \textbf{SFT with Compressed Reasoning Data}: We include baselines trained via supervised fine-tuning on datasets augmented with compressed reasoning chains, such as CoT-Valve~\cite{ma2025cot} and LS-Mixture~\cite{yu2025long}, to assess the benefit of incorporating distilled reasoning steps.  

\subsection{Comparison with Baselines}

\begin{figure*}[tb]
    \centering
    \includegraphics[width=\textwidth]{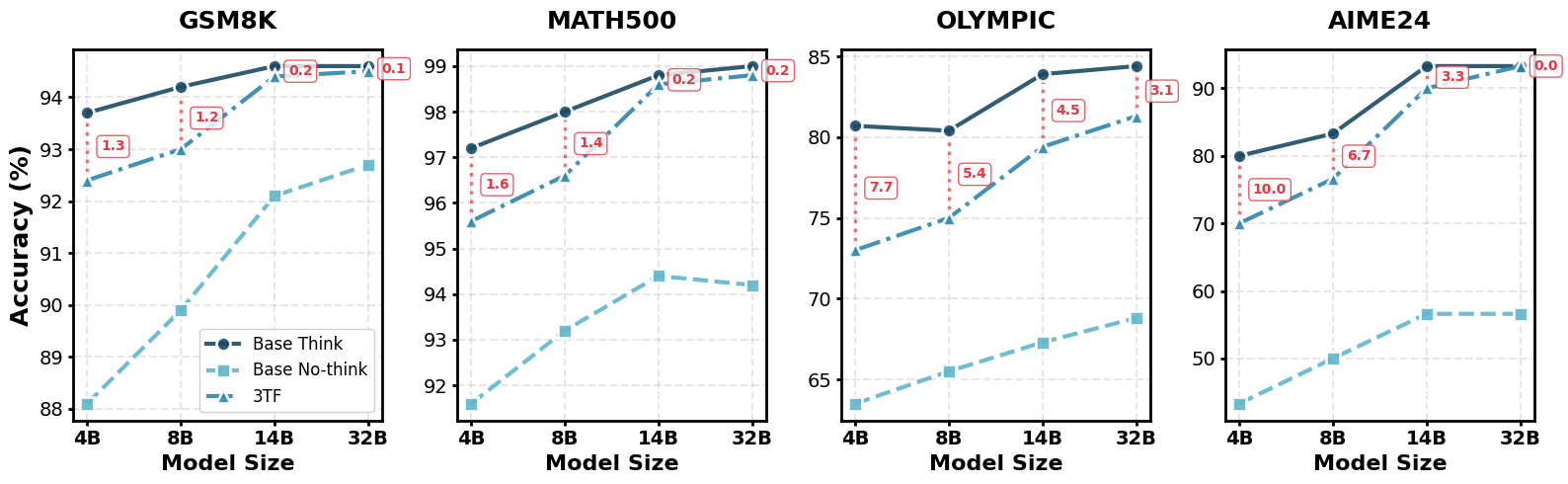}
    \caption{Accuracy scaling of 3TF from 4B to 32B across four reasoning benchmarks. As model size increases, 3TF progressively closes the gap to the \texttt{Base-Think} mode, showing improved reasoning recoverability at scale.}
\label{fig:scaling}
\end{figure*}

\begin{figure*}[tb]
    \centering
    \includegraphics[width=\textwidth]{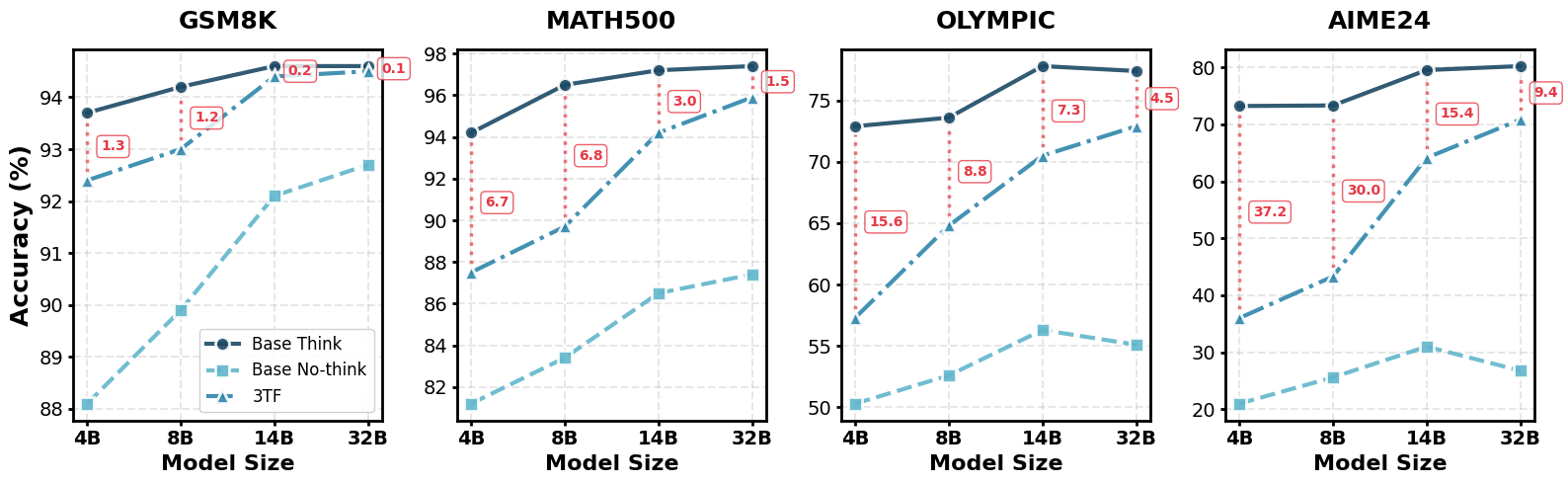}
    \caption{Avg@k performance scaling of 3TF from 4B to 32B across four reasoning benchmarks. Similar to the pass@k accuracy results, 3TF progressively closes the gap to the \texttt{Base-Think} mode, showing that reasoning recoverability for this metric also improves with scale.}
\label{fig:scaling_pass1}
\end{figure*}

\begin{figure*}[tb]
    \centering
    \includegraphics[width=\textwidth]{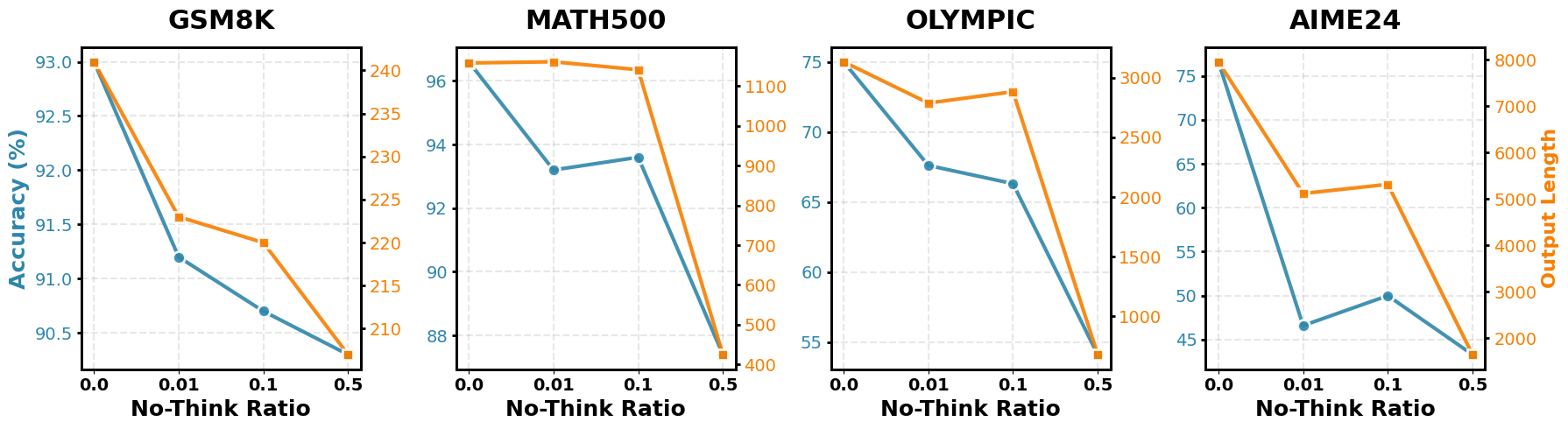}
    \caption{Impact of mixing No-Think data during training on Qwen3-8B across four reasoning benchmarks. Increasing the No-Think ratio consistently shortens model outputs (orange) but monotonically reduces accuracy (blue). Even 1\% No-Think materially hurts reasoning fidelity, indicating high sensitivity to non-reasoning supervision.}
    \label{fig:no-think}
\end{figure*}

Table~\ref{tab:main} reports the results of 3TF and all baselines across four reasoning benchmarks and three model scales. Overall, 3TF achieves a strong trade-off between reasoning accuracy and output efficiency.

\paragraph{Overall Performance.}
Across all datasets, 3TF attains accuracy comparable to the full Thinking mode while producing substantially shorter responses. 
For instance, on Qwen3-14B, 3TF achieves 94.4\% on GSM8K and 97.8\% on MATH500, nearly matching the full-CoT performance but using only 19\% and 43\% of the output tokens on GSM8K and MATH500, respectively. 
Similar trends hold across smaller model variants, showing that the proposed asymmetric training--inference design effectively preserves reasoning ability without explicit reasoning traces. 
3TF further adapts its output length to task complexity, remaining highly concise on easier datasets while allowing moderately longer responses on more challenging ones.

\paragraph{Comparison.}
Prompt-based methods generally maintain the original model performance in terms of accuracy, but achieve lower compression ratios compared to 3TF. 
This gap is particularly evident on simpler datasets such as GSM8K and MATH500, where 3TF produces much more compact outputs. 
For fine-tuning based methods, CoT-Valve achieves high compression but suffers a significant drop in performance, whereas the model trained with LM-Mixture maintains accuracy but yields a much lower compression ratio. 
Taken together, 3TF achieves a strong trade-off between accuracy and output compression.

\begin{tcolorbox}[title=\textbf{Takeaway \#1}]
Training on reasoning-rich data alone can markedly improve the performance of non-thinking inference modes while keeping outputs far more concise than full reasoning traces.
\end{tcolorbox}

\subsection{Scaling Law of Reasoning Recoverability}

As shown in Figure~\ref{fig:scaling}, we analyze the scaling behavior of 3TF across model sizes from 4B to 32B parameters. A clear monotonic trend emerges: as the parameter count increases, the performance gap between 3TF and the Base\_Think mode consistently narrows across all four benchmarks. On GSM8K and MATH500, the trained No-Think model at 14B already nearly saturates full CoT performance, and at 32B the remaining delta is minimal. Even on more challenging distributions such as Olympic and AIME24, where smaller models still show non-trivial degradation, the 32B checkpoint closes most of the gap.

These results highlight a key scaling dynamic: larger reasoning-centric LMs have stronger latent reasoning priors that can be efficiently unlocked via asymmetric 3TF training. As scale increases, reasoning ability becomes more accessible without explicit chain-of-thought.

\begin{tcolorbox}[title=Takeaway \#2, breakable]
Scaling up the model steadily narrows the gap between 3TF and full Thinking mode, bringing larger models closer to full reasoning performance.
\end{tcolorbox}

\subsection{Avg@k Results}

As shown in Figure~\ref{fig:scaling_pass1}, 3TF demonstrates a consistent scaling trend under Avg@k. Across all four benchmarks, performance improves monotonically with model size, narrowing the gap to Base-Think. On GSM8K and MATH500, 3TF at 32B nearly matches full CoT, with differences under 1--2\%, suggesting that larger models recover most of the reasoning benefits even without explicit chain-of-thought during inference.

In more challenging settings (OLYMPIC, AIME24), smaller models show significant performance degradation compared to Base-Think (up to 30--37\% at 4B). However, scaling mitigates this gap: at 32B, 3TF recovers over 10\% on OLYMPIC and more than 25\% on AIME24, highlighting the strong recoverability of reasoning through scale. 

\begin{tcolorbox}[title=\textbf{Takeaway \#3}]
For smaller models, 3TF under Avg@k lags behind the Thinking mode, but this gap steadily diminishes as model scale increases.
\end{tcolorbox}

\subsection{Effect of Adding No-Think Data}

We further evaluate the impact of training with No-Think data on Qwen3-8B. Adding No-Think samples consistently shortens outputs across all datasets (e.g., AIME24 from 7961 to 1642). However, this compression is coupled with substantial accuracy loss: GSM8K drops from 93.0\% to 90.3\%, and AIME24 from 76.6\% to 43.3\%. MATH500 and OLYMPIC exhibit the same trend. Figure~\ref{fig:no-think} visualizes the effect of different No-Think ratios. Even extremely small amounts (1\% of training) cause noticeable degradation across all benchmarks. Thus, while No-Think data reliably enforces conciseness, we hypothesize that the inclusion of non-thought data causes the model to revert to a purely non-reasoning mode, which undermines its ability to perform complex reasoning tasks and leads to performance degradation.

\begin{tcolorbox}[title=\textbf{Takeaway \#4}]
Adding No-Think data significantly reduces 3TF performance.
\end{tcolorbox}

\subsection{Ablation Study on the Datasets}

\begin{table}[ht]
\centering
\setlength{\tabcolsep}{1.6mm}
\renewcommand{\arraystretch}{1.0}
\begin{tabular}{lcccc}
\toprule
\multirow{2}{*}{\textbf{Methods}} & \multicolumn{2}{c}{\textbf{GSM8K@1}} & \multicolumn{2}{c}{\textbf{MATH-500@4}} \\
\cmidrule(ll){2-3} \cmidrule(ll){4-5}
& Acc. & Tok. & Acc. & Tok. \\
\midrule
\rowcolor{lightgray!30}\multicolumn{5}{c}{\textbf{\textit{Qwen3-8B}}} \\
\midrule
NoThink & 89.9 & 245 & 93.2 & 1092 \\
NoThink-T1 & 90.0 & 300 & 91.2 & 741 \\
NoThink-T2 & 90.2 & 250 & 93.4 & 1038 \\
\rowcolor{c0!5}
\textbf{3TF} & 93.0 & 241 & 96.6 & 1358 \\
\bottomrule
\end{tabular}
\caption{Ablation study on Qwen3-8B comparing different training methods: NoThink, NoThink-T1, NoThink-T2, and 3TF.}

\label{tab:ablation_data}
\end{table}

In this section, we compare 3TF with two training strategies: NoThink-T1, which uses thought data but excludes reasoning part, and NoThink-T2, which is trained from distilled non-thought data (from Qwen3 Non-Think mode). While NoThink-T1 results in slightly shorter outputs, its performance suffers noticeably compared to 3TF. On the other hand, NoThink-T2, trained directly from non-thought data, shows performance comparable to the baseline non-thought model, with no significant improvements. These results suggest that 3TF effectively combines reasoning supervision with thought-free inference, outperforming methods that prioritize direct fine-tuning with non-thought data.

\begin{tcolorbox}[title=\textbf{Takeaway \#5}]
Data containing reasoning improves the performance of No-Think modes, rather than the answer content alone.
\end{tcolorbox}

\subsection{Train from Scratch}

\begin{table*}[tb]
\centering
\setlength{\tabcolsep}{1.6mm}
\renewcommand{\arraystretch}{1.05}
\begin{tabular}{lcccccccc}
\toprule
\multirow{2}{*}{\textbf{Methods}} & \multicolumn{2}{c}{\textbf{GSM8K@1}} & \multicolumn{2}{c}{\textbf{MATH-500@4}} & \multicolumn{2}{c}{\textbf{OLYMPIC@4}} & \multicolumn{2}{c}{\textbf{AIME 2024@16}} \\
\cmidrule(lr){2-3} \cmidrule(lr){4-5} \cmidrule(lr){6-7} \cmidrule(lr){8-9}
&Acc. &Tok. &Acc. &Tok. &Acc. &Tok. &Acc. &Tok.\\
\midrule
\rowcolor{lightgray!30}\multicolumn{9}{c}{\textbf{\textit{Qwen2.5-7B-Instruct}}} \\
\midrule
NoThinking & 80.8 & 278 & 86.6 & 576 & 50.0 & 935 & 26.6 & 940\\
Hybrid (Think) & 92.6 & 1834 & 94.0 & 7352 & 67.2 & 13241 & 56.6 & 20331\\
Hybrid (NoThink) & 88.4 & 310 & 90.0 & 1727 & 56.5 & 2253 & 46.6 & 5023\\
\rowcolor{c0!5}
\textbf{3TF} & 91.0 & 589 & 94.6 & 4143 & 63.0 & 5260 & 56.6 & 12025\\
\midrule
\rowcolor{lightgray!30}\multicolumn{9}{c}{\textbf{\textit{Qwen2.5-7B-R1-Distill}}} \\
\midrule
Think & 86.4 & 541 & 96.4 & 2971 & 77.6 & 7690 & 76.6 & 10414\\
Hybrid (Think) & 87.0 & 322 & 96.0 & 3601 & 68.2 & 2410 & 76.6 & 11786\\
Hybrid (NoThink) & 86.8 & 271 & 90.0 & 778 & 61.8 & 1200 & 43.3 & 2654\\
\rowcolor{c0!5}
\textbf{3TF} & 88.6 & 287 & 93.6 & 913 & 64.2 & 2008 & 60.0 & 4346\\
\bottomrule
\end{tabular}
\caption{Results of 3TF applied to Qwen2.5-7B models first trained on mixed \texttt{Think}/\texttt{NoThink} data to form hybrid reasoning, then fine-tuned with reasoning-only samples.}

\label{tab:qwen25_7b}
\end{table*}

\begin{table}[htb]
\centering
\setlength{\tabcolsep}{2pt} 
\begin{tabular}{@{}lrrrr@{}}
\toprule
\textbf{} & \textbf{GSM8K} & \textbf{MATH} & \textbf{Olympic} & \textbf{AIME} \\
\textbf{} & \textbf{(1319)} & \textbf{(500)} & \textbf{(674)} & \textbf{(30)} \\
\midrule
\multicolumn{5}{l}{\textit{Total Count}} \\
\midrule
Think    & 61,936 & 42,722 & 140,563 & 6,903 \\
NoThink  &    485 &    959 &   2,134 &   137 \\
3TF      &  1,822 &  1,746 &  13,120 &   1,032 \\
\midrule
\multicolumn{5}{l}{\textit{Average Count per Question}} \\
\midrule
Think    & 46.96  & 85.44  & 208.55  & 230.10 \\
NoThink  &  0.37  &  1.92  &   3.17  &   4.57 \\
3TF      &  1.38  &  3.49  &  19.47  &  34.40 \\
\bottomrule
\end{tabular}
\caption{Total count and average count per question of reasoning tokens across four benchmarks under Think, NoThink, and 3TF. Numbers in parentheses indicate the number of questions in each benchmark.}
\label{tab:total_count}
\end{table}
Unlike previous experiments where 3TF was fine-tuned directly on hybrid reasoning models like Qwen3, we first train a base model into a hybrid reasoning form using mixed Think and NoThink data. We then apply 3TF training solely on reasoning samples. This setup tests whether asymmetric supervision can be effective without relying on pre-existing hybrid priors.

As shown in Table~\ref{tab:qwen25_7b}, 3TF consistently achieves high accuracy with concise outputs. On Qwen2.5-7B-Instruct, it reaches 91.0\% on GSM8K and 94.6\% on MATH500, comparable to the Hybrid (Think) model while using only about a third of its tokens. Similar improvements are observed on Qwen2.5-7B-R1-Distill, where 3TF achieves 88.6\% and 93.6\% accuracy on GSM8K and MATH500 with compact responses. The distilled variant shows stronger reasoning recoverability; on AIME24, it reaches 60.0\% accuracy with less than half the token length of the Instruct version. Overall, 3TF proves effective even when trained from scratch, demonstrating that asymmetric reasoning fine-tuning can independently induce concise yet faithful reasoning behavior.

\subsection{Analysis of Output Structure}

To understand how 3TF alters the internal structure of model outputs, we compare the frequency of canonical "reasoning words" across Think, No-Think, and 3TF. These reasoning proxies (e.g., "wait", "however", "check") are listed in the Appendix, with Table~\ref{tab:total_count} reporting both total counts and per-question frequencies across four reasoning benchmarks. 3TF increases the use of reasoning markers relative to No-Think but remains far below Think, often by more than an order of magnitude. This suggests that 3TF does not fully revert to chain-of-thought reasoning, instead maintaining concise, targeted inference traces for efficiency. As benchmark difficulty increases (e.g., Olympic, AIME), all modes, including 3TF, emit more reasoning markers and generate longer outputs, reflecting adaptive reasoning exposure based on task complexity. Thus, 3TF induces a sparse reasoning regime: it is not pure No-Think but is significantly shorter than Think, only surfacing critical reasoning when needed.

\begin{tcolorbox}[title=\textbf{Takeaway \#6}]
3TF introduces limited reasoning into non-thinking mode and adapts: the harder the task, the more reasoning patterns are expressed.
\end{tcolorbox}

\section{Conclusion}\label{sec:con}
In this work, we presented \textbf{3TF}, a novel framework for efficient reasoning in large language models. By first training a hybrid model and then further training on chain-of-thought annotated data while enforcing concise, thought-free outputs at inference in the no-reasoning mode, 3TF enables models to internalize rich reasoning patterns while avoiding verbose explanations. Extensive experiments across multiple benchmarks and model scales demonstrate that 3TF preserves reasoning accuracy, reduces output length, and adapts to problem difficulty. Our results highlight a practical approach for scalable and efficient implicit reasoning, offering a new direction for deploying reasoning-capable LLMs with minimal inference overhead.

\clearpage
\section{Limitations}\label{sec:limitations}

While 3TF demonstrates strong performance and efficiency across multiple benchmarks and model scales, there are several limitations worth noting. First, on smaller-scale models (e.g., Qwen3-4B), the \texttt{pass@1} accuracy remains notably lower compared to larger models, indicating that reasoning recoverability under minimal output sampling is still limited at small scale. Second, although 3TF reduces the number of generated tokens, it may still produce slightly longer outputs on particularly challenging benchmarks (e.g., AIME24 and Olympiad) to preserve accuracy, which partially offsets efficiency gains. Finally, 3TF has been evaluated primarily on mathematics-focused benchmarks, and its generalizability to other domains remains to be thoroughly investigated. Addressing these limitations is an important direction for future work, especially for applying 3TF to smaller models or broader domains.


\bibliography{main}

\clearpage

\appendix

\section{Appendix}

\subsection{Experimental environment}
We trained on the NVIDIA 8*H200 node, with CUDA version 12.7 and CPU model Intel(R) Xeon(R) Platinum 8468V.

\subsection{Reasoning Words}

The table below lists the reasoning words we use:
\begin{table}[h]
    \centering
    \renewcommand{\arraystretch}{1.2}
    \begin{tabular}{@{}c@{}c@{}}
        \toprule
        \textbf{Reasoning Words} \\
        \midrule
        ``wait'', ``alternatively'', ``but'', ``however'',\\ 
        ``alternative'', ``check'', ``double-check'', \\
        ``hmm'', ``okay'', ``maybe''\\
        \bottomrule
    \end{tabular}
\end{table}
\subsection{Prompt}

The prompt used in the Prompt method is shown below.

\begin{tcolorbox}[title=CoD] 
Think step by step, but only keep a minimum draft for each thinking step, with 5 words at most.
\end{tcolorbox}

\begin{tcolorbox}[title=CCoT] 
be concise
\end{tcolorbox}
\end{document}